\documentclass[11pt,a4paper]{article}
\usepackage[hyperref]{naaclhlt2019}
\usepackage{times}
\usepackage{latexsym}

\usepackage{url}
\usepackage{graphicx}
\usepackage{amssymb,amsmath, bm, xcolor}
\usepackage{float} 
\usepackage{subcaption} 
\graphicspath{.}

\aclfinalcopy 

\usepackage[symbol]{footmisc}

\title{A Bi-Encoder LSTM Model For Learning Unstructured Dialogs}

\author{Diwanshu Shekhar\footnotemark \\
  University of Denver \\\And
  Pooran S. Negi\footnotemark \\
  University of Denver\\\And
  Mohammand Mahoor\footnotemark\\
  University of Denver \\
  }

\date{}

\begin{document}
\maketitle
\footnotetext[1]{diwanshu.shekhar@du.edu}
\footnotetext[2]{pooran.negi@du.edu}
\footnotetext[3]{mohammad.mahoor@du.edu}
\begin{abstract}
 Creating a data-driven model that is trained on a large dataset of unstructured dialogs is a crucial step in developing Retrieval-based Chatbot systems. This paper presents a Long Short Term Memory (LSTM) based architecture that learns unstructured multi-turn dialogs and provides results on the task of selecting the best response from a collection of given responses.  Ubuntu Dialog Corpus Version 2 was used as the corpus for training. We show that our model achieves 0.8\%, 1.0\% and 0.3\% higher accuracy for Recall@1, Recall@2 and Recall@5 respectively than the benchmark model. We also show results on experiments performed by using several similarity functions, model hyper-parameters and word embeddings on the proposed architecture

\end{abstract}

\section{Introduction}

Recently statistical techniques based on recurrent neural networks (RNN) have achieved  remarkable successes in a variety of natural language processing tasks, leading to a great deal of commercial and academic interests in the field \cite{Yoshua:13,Eric:14}. Significant progress in the area of Machine Translation, Text Categorization, Spam Filtering, and Summarization have been made. Research in developing Dialog Systems or Conversational Agents - perhaps a desirable application of the future- have been growing rapidly. A Dialog System can communicate with human in text, speech or both and can be classified into - Task-oriented Systems and Chatbot Systems. 

Task-oriented systems are designed for a particular task and set up to have short conversations. These systems interact with humans to get information to help complete the task. These include the digital assistants that are now on every cellphone or on home controllers and voice assistants such as Siri, Cortana, Alexa, Google Now/Home, etc.

Chatbot Systems, the  area of this paper, are systems that can carry on extended conversations with the goal of mimicking unstructured conversations or ‘chats’ characteristic of human-human interaction. \newcite{Ryan:15} explored learning models such as TF-IDF (Term Frequency-Inverse Document Frequency),  Recurrent Neural Network (RNN) and a Dual Encoder (DE) based on Long Short Term Memory (LSTM) model  suitable  to  learn  from  the  Ubuntu  Dialog  Corpus  Version 1 (UDCv1). We use this same architecture but on Ubuntu Dialog Corpus Version 2 (UDCv2) as a benchmark and introduce a new LSTM based architecture called the Bi-Encoder LSTM model (BE) that achieves 0.8\%, 1.0\% and 0.3\% higher accuracy for Recall@1, Recall@2 and Recall@5 respectively than the DE model.  In contrast  to  the  DE  model, the  proposed BE  model  has  separate  LSTM networks for encoding utterances  and responses.  The BE model also has a different similarity measure for utterance and response matching than that of the benchmark model. We further show results of various experiments necessary to select the best similarity function, hyper-parameters and word embedding for the BE model.

Section 2 describes the related current state-of-the-art research on Chatbot Systems. We describe the proposed BE model in Section 3. The experiments and results are described in Section 4 and, we conclude the paper in Section 5 with suggestions for potential future work.

\section{Background}
For clarity, we establish a notation in this paper wherein the type of the mathematical quantity involved will be denoted by its representation. Scalars are represented by lower case letters  $i, j, k, \cdots; \alpha, \beta, \gamma, \cdots$, vectors are represented via lower case bold letters $\bm{a,b, \cdots, e, \epsilon}, \cdots$ and matrices are represented by bold upper case letters $\bm{A,B}, \cdots, \bm{E}, \cdots$. Calligraphic letters $\mathcal{A, T, \cdots}$ are used to represent sets of objects. We consistently follow similar convention for functions where $f$ represents scalar valued functions, bold $\bm{f}$ represents vector valued functions and bold $\bm{A_{i,.}}$ represents the $i^{th}$ row of the matrix $A$. 
\subsection{Related Work}
Early Chatbot systems such as ELIZA \cite{Joseph:66}, ALICE \cite{Richard:08} and PARRY \cite{Kenneth:71} were based on pattern matching where a human statement was matched to a pattern and a response was retrieved that pertained to the matched pattern. These Chatbot systems were rule based and needed domain expertise to hand-craft rules in advance which made the design of these systems very expensive and tedious. To address this limitation, the idea of corpora-based Chatbot System was introduced. At the time of this research, two large corpora are available to design the corpora-based Chatbot Systems - Twitter Corpus \cite{Alan:10} and the Ubuntu Dialog Corpus \cite{Ryan:15}. \newcite{Iulian:15} did a survey of all available corpora for corpora-based Chatbot systems.

A type of corpora-based Chatbot systems that has been popular is the Information Retrieval (IR) based Chatbot systems. In the IR-based Chatbot systems, an utterance is matched to a repository of responses and the response that matches the most is retrieved. If this repository is too big, the retrieval process may be too slow. To address this problem, \newcite{Sina:10} devised a filtering technique based on feature selection to reduce the size of the set of responses to match the given utterance with. \newcite{Hao:13} used the same filtering technique but used RankSVM to match utterance with responses. \newcite{Ryan:15} used LSTM-based Dual Encoder model (DE) to retrieve the best response from a set of responses of size 10 (since the set of responses to choose from was already small filtering was not necessary). \newcite{Rudolf:15} showed that an ensemble of LSTM, DE and CNN models performed better than the DE model. 

Another type of corpora-based Chatbot system is the Generative Chatbot system. One clear benefit of the Generative systems is that they don't need a repository of responses to choose from as a response is generated by the system itself.  \newcite{Alan:11} used Sequence-to-Sequence RNN (seq2seq) model, a model that is commonly used for Machine Translation \cite{Ilya:14}, to generate a response given an utterance. Although seq2seq models works really well in Machine translation, the model did not perform very well in the response generation task as in machine translation words or phrases in the source and target sentences tend to align well with each other; but in dialogs, a user utterance may share no words or phrases with a coherent response. Several modifications of seq2seq model have been made for response generation. \newcite{Jiwei:16a} made modification to address the problem of seq2seq model producing responses like ``I'm OK'' or ``I don't know'' that tend to end the conversation. \newcite{Ryan:17} used hierarchical approach to use longer prior context in the seq2seq model. The basic seq2seq model focuses on generating single responses, and so don't tend to do a good job of continuously generating responses that cohere across multiple turns. This can be addressed by using reinforcement learning \cite{Jiwei:16b}, as well as techniques like adversarial networks \cite{Jiwei:17} that can select multiple responses that make the overall conversation more natural.

Not all Generative Chatbot systems are based on seq2seq model. \newcite{Lifeng:15} showed that transduction models can be used to generate response. \newcite{Wen:15} presented a statistical language generator based on a semantically controlled Long Short-term Memory (LSTM) structure. Although not related to Chatbot systems, \cite{Mei:16} introduced an LSTM-E architecture that was able to generate a description given a video. \cite{Anjuli:16} demonstrated a hybrid system called Smart Reply that leverages both the Retrieval and Generative concepts. At the time of this research, the generative-based systems are not doing so well and most production systems are essentially retrieval-based such as -  Cleverbot \cite{Rollo:16} and Microsoft's Little Bing system.

\section{Bi Encoder Model}

\begin{figure}
    \centering
    \includegraphics[scale=0.35]{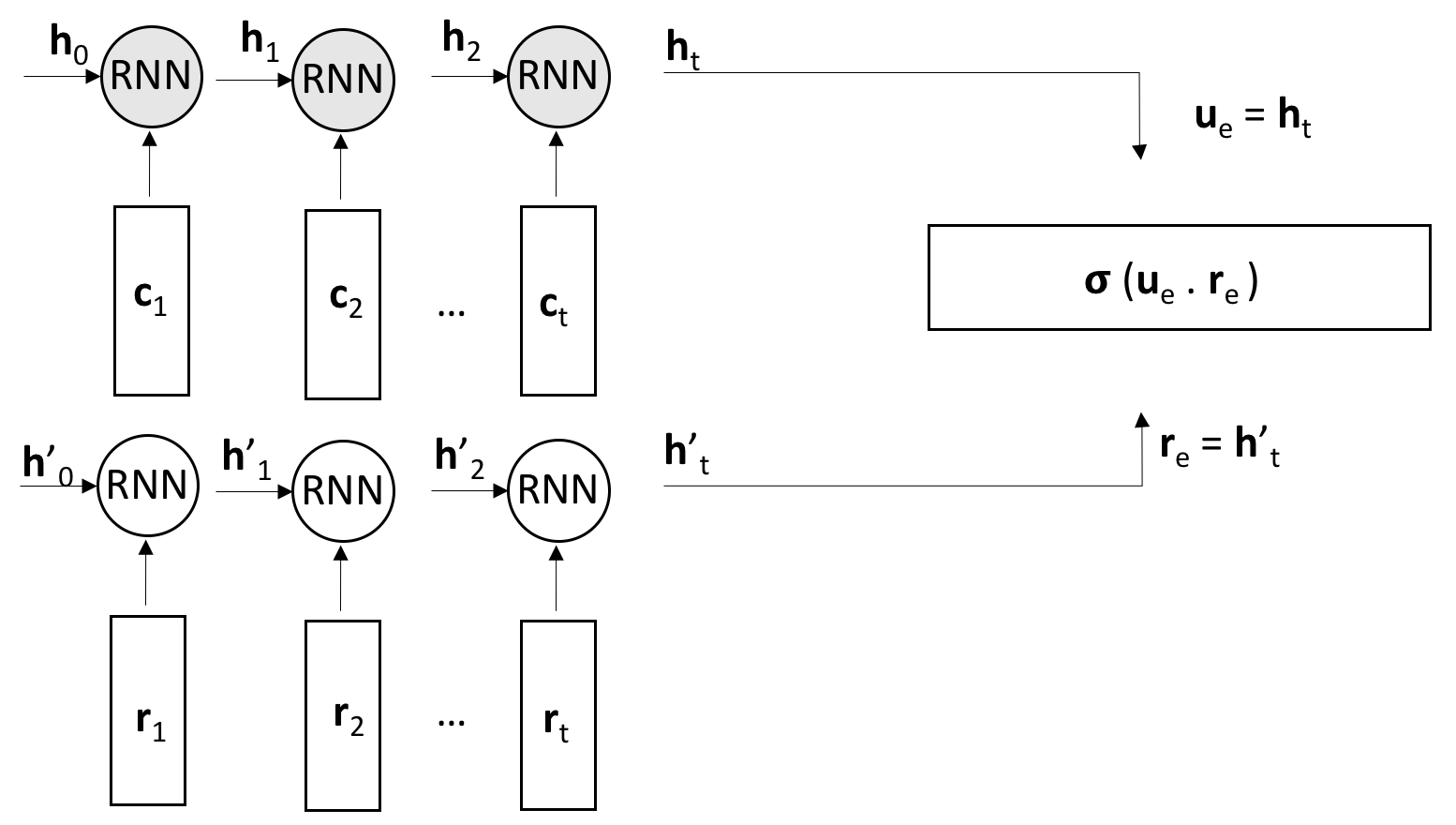}
    \caption{Bi Encoder LSTM Architecture. RNNs are colored in grey and white to show two different LSTM networks}
    \label{fig:BE}
\end{figure}

The proposed BE model architecture in Figure~\ref{fig:BE} is motivated by the typical setup of conversation between two persons. Each person has to encode long and short term conversation contexts to best respond to a spoken sentence (an utterance or context).

As a natural design choice, in the BE model one LSTM cell (colored in grey) learns encoding of an utterance (or questions or contexts) while the other LSTM cell (colored in white) learn encoding of a response (or answers or responses). A sequence of GloVe embedding vectors \cite{Jeffrey:14} of an utterance are fed into the upper LSTM while the sequence of embedding vectors of a response are fed into the lower LSTM cell. Vectors representing the final states \(\bm{h}_t \in \mathbb{R}^{s} \) of the upper and lower LSTM cells  are used for final representations of the utterance and the response as $\bm{u}_e, \,\bm{r}_e$ respectively. To drive learning of $\bm{u}_e$ and $\bm{r}_e$,  we measure their similarity in the hidden vector space using dot product i.e

\begin{align}
\text{sim}(\bm{u}_e, \bm{r}_e) = \langle \bm{u}_e^T,  \bm{r}_e \rangle \label{eq.sim}
\end{align}

The training of the model via BPTT is done by minimizing the binary cross entropy  $\mathcal{X}(q, \, p) $  between the learned probability $p$ 
and the ground truth paring probability $q = \{0,\, 1\}$, where $1$ denotes 
$\bm{u_e}, \bm{r_e}$ are genuinely paired and $0$ denotes otherwise. 
Using similarity in Eq. \eqref{eq.sim}, $p$ is calculated as:

\begin{align}
p = \frac{1}{1 + e^{(\text{sim}(\bm{u}_e, \bm{r}_e) + \bm{b})}} \label{eq.prob}
\end{align}

In eq. \eqref{eq.prob}, $b$ is a scalar free parameter bias that is learned by the model.
The aforementioned cross entropy loss is given by:
\begin{align}
\mathcal{X}(q, \, p) = - q \cdot log(p) - (1 - q) log(1- p) \label{eq.loss}
\end{align}

The model is trained using the Adam Optimizer \cite{Diederik:17} with a learning rate of $0.001$ by minimizing the loss function in Eq. \eqref{eq.loss}.

\begin{figure}[H]
    \centering
    \includegraphics[scale=0.55]{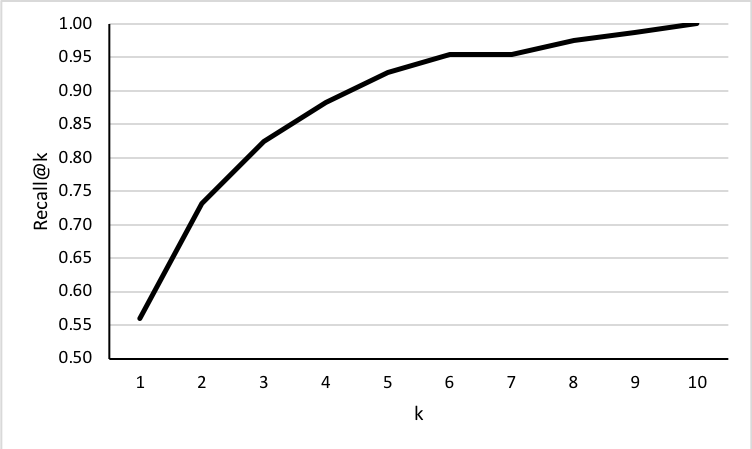}
    \caption{CMC Curve of the BE Model}
    \label{fig:cmc}
\end{figure}

Figure~\ref{fig:cmc} shows the Cumulative Match Characteristic (CMC) curve that shows the true positive identification rate of the BE model for Recall@k for k $\in \{1, 2, ..., 10\}$.

In this subsequent section, we look at various experiments that helped us decide to select the best similarity function, hyper-parameters and word embedding for the BE model. We also show performance of the BE model in comparison to the DE model.

\section{Experiments and Results}
All our models were implemented in Tensorflow v1.7 and trained using a GeForce GTX 1080 Ti NVIDIA GPU. We used the same training data, UDC as \cite{Ryan:15} but its second version (UDCv2). Models are trained on 1 million pairs of utterances and responses on the training set and evaluated against a test set. We fine-tune the model with hyper-parameters, determine the optimum similarity function and word embedding using the validation dataset. 

For evaluation and  model selection, we present our model with 10 response candidates, consisting of one right response and the rest nine incorrect responses. This set of 10 response candidates per context is provided in the validation and test set in UDCv2 (more details in Section~\ref{sec:data}). The model ranks these responses and its ranking is considered correct if the correct response is among the first $k$ candidates. This quantity is denoted as Recall@k. Specifically, we report mean values of Recall@1, Recall@2 and Recall@5.

\begin{figure}
    \centering
    \includegraphics[scale=0.4]{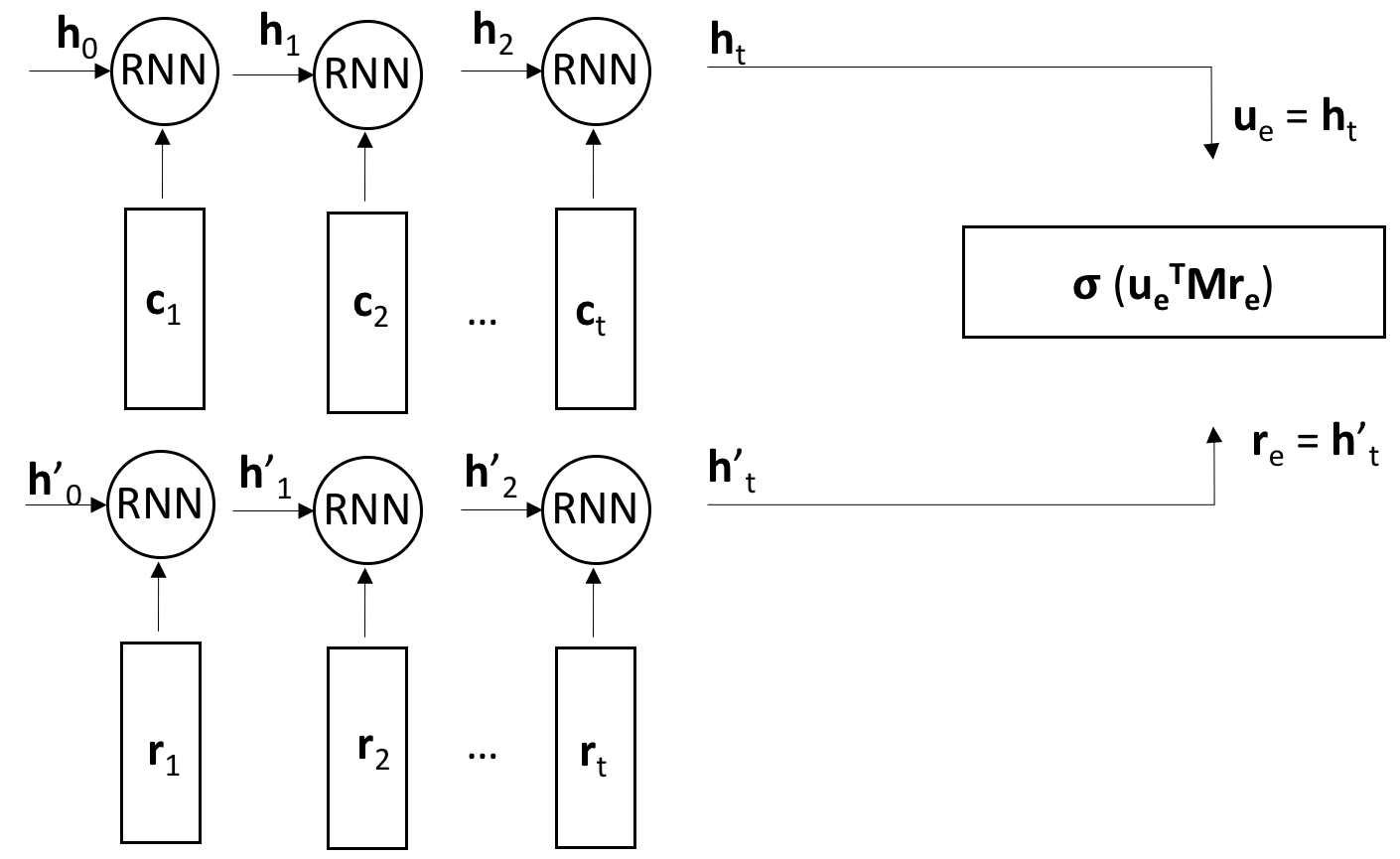}
    \caption{DE Model. All RNNs are colored in white to show the same LSTM network has been fed first by the utterance and then by the response}
    \label{fig:DE}
\end{figure}

For benchmarking, we use the DE model in \cite{Ryan:15} and the results of the DE model on UDCv2 as published in \\ \url{https://github.com/rkadlec/ubuntu-ranking-dataset-creator}. In contrast to the BE model, the DE model has one LSTM cell that encodes both the utterance and the response. The encoding for the utterance, $\bm{u}_e$ is multiplied with a trainable matrix $\bm{M}$ whose result is compared with the encoding for response, $\bm{r}_e$ by a dot product (Figure~\ref{fig:DE}).

We also reproduced the DE model for comparison and we refer it as the DER model. Note that the DE model was originally modeled and trained in Theano. 

\subsection{Data}\label{sec:data}
The Ubuntu Dialog Corpus (UDC) is the largest freely available multi-turn dialog corpus \cite{Ryan:15}. It was constructed from the Ubuntu chat logs - a collection of logs from Ubuntu-related chat rooms on the Freenode IRC network. Although multiple users can talk at the same time in the chat room, the logs were pre-processed using heuristics to create two-person conversations. The resulting corpus consists of almost one million two-person conversations, where a user seeks help with his/her Ubuntu-related problems (the average length of a dialog is 8 turns, with a minimum of 3 turns). Because of its size, the corpus is well-suited for deep learning in the context of dialogue systems. 

UDCv2 released in 2017 made several significant updates to its predecesor (\url{https://github.com/rkadlec/ubuntu-ranking-dataset-creator}). To summarize - UDCv2 is separated into training, validation and test set; the sampling procedure for the context length in the validation and test set is changed from an inverse distribution to a uniform distribution; the tokenization and entity replacement procedure was removed; differentiation between the end of an utterance (\_\_eou\_\_) and end of turn (\_\_eot\_\_) has been added; a bug that caused the distribution of false responses in the test and validation sets to be different from the true responses was fixed.

The training set consists of labelled 1 million pairs of utterances and responses. It has equal distribution of true context-response pairs labeled as $1$ versus the context-distraction pairs labeled as $0$. Keeping all the words that occur at least 5 times, the training set has a vocabulary of 91,620. The average utterance is 86 words long and the average response is 17 words long.

\begin{table}
\begin{center}
\begin{tabular}{|p{2.5cm}|p{2.5cm}|p{0.8cm}|}
\hline \bf Context & \bf Response & \bf Label \\
\hline
you just click userprefer and creat one ... \_\_eou\_\_ \_\_eot\_\_ i ca n't access the wiki at all - it throw http auth at me \_\_eou\_\_ \_\_eot\_\_ & unless nat get distract by someth shinier & 1 \\
\hline
i think that tutori be outdat - veri first instruct fail \_\_eou\_\_ and it differ slight to what ubotu have to say \_\_eou\_\_ & i think that tutori be outdat - veri first instruct fail \_\_eou\_\_ and it differ slight to what ubotu have to say \_\_eou\_\_ & 0 \\
\hline
\end{tabular}
\end{center}
\caption{\label{train-sample} Examples from the training dataset of UDCv2 showing both the correct (1) and incorrect response (0) labels}
\end{table}

The validation dataset consists of 19,560 examples where each example consists of a context and 10 responses where the first response is always the true response. The test dataset, structured the same as the validation dataset, consists of 18920 examples. The correct response is
the actual next utterance in the dialogue and a false response is randomly sampled utterance from elsewhere within a set of dialogues in UDC that has been set aside for creation of validation and test set \cite{Ryan:15}. The words of the UDCv2 are stemmed (strip suffixes from the end of the word), and lemmatized (normalize words that have the same root, despite their surface differences).

\subsection{Effect of similarity measures}
In the BE model, we used dot product similarity between the encoded utterance $\bm{u}_t$ and response $\bm{r}_e$. Before we made that decision, we evaluted severa other similarity measures. The description of these similarity measures are given in the subsequent sections.

\subsubsection{Cosine Similarity}
Instead of taking the dot product of \(\bm{u}_e\) and \(\bm{r}_e\), we ignore their magnitude and take the dot product of their unit vectors. This is shown in the following equation:

\begin{align} \label{eq:cosim}
\text{sim}(\bm{u}_e, \bm{r}_e) = \frac{\bm{u}_e^T \cdot \bm{r}_e}{\left | \bm{u}_e \right |\left | \bm{r}_e \right |}
\end{align}

\subsubsection{Polynomial Similarity}
In machine learning, the polynomial kernel is a kernel function commonly used with support vector machines (SVMs) and other kernelized models. Although the Radial Basis Function (RBF) kernel is more popular in SVM classification than the polynomial kernel, \newcite{Yoav:08} showed that polynomial kernel gives better result than the RBF-Kernel for NLP applications: 

For degree$-d$ polynomials, the polynomial kernel is defined as:
\begin{align} \label{eq:polsim}
K(\bm{x},\bm{y}) = (\bm{x}^T \cdot \bm{y} + c)^d
\end{align}
where $x$ and $y$ are vectors in the input space, i.e. vectors of features computed from training or test samples and $c$ \(\geq 0\) is a free parameter trading off the influence of higher-order versus lower-order terms in the polynomial. When $c = 0$, the kernel is called homogeneous.

In this experiment, we used the polynomial kernel from \(0^{th}\) to the \(3^{rd}\) degree for the similarity measure. The following equation gives the similarity function:
\begin{align}
\text{sim}(\bm{u}_e, \bm{r}_e) = \sum_{d=0}^{3}(\bm{u}_e^T\cdot \bm{r}_e)^d
\end{align}

\subsection{Effect of using all hidden states}
\label{ssec:layout}
In this experiment, we used all the hidden states of the LSTM and not just the final states to encode the utterance and response. The encoding for the utterance is given by:
\begin{align} \label{eq:hiddenu}
\bm{u_e} = \sum_{t=1}^{T}\frac{t^2}{T^2}\cdot \bm{h}_t
\end{align}
where $T$ is the maximum context length of the utterance

Similarly, the encoding for the response is given by:
\begin{align} \label{eq:hiddenr}
\bm{r_e} = \sum_{t=1}^{T}\frac{t^2}{T^2}\cdot \bm{h'}_t
\end{align}

We keep the similarity function as in the original BE model as shown in Eq. \eqref{eq.sim}.
         
\subsection{Deep LSTM}

In this experiment, we added two more layers to the shallow LSTM BE model and looked at the result. We keep the similarity function as in the original BE model as shown in Eq. \eqref{eq.sim}.

\subsection{Results and Discussion}

Table~\ref{BE-PER-table} compares the performance of the proposed BE model, the benchmark DE model and the reproduced DE model, the DER model on UDCv2 dataset. 
Compared to benchmark DE model, the proposed BE model achieves $0.8\%$, $1.0\%$ and $0.3\%$ higher accuracy for Recall@1, Recall@2 and Recall@5 respectively. Note that compared to the reproduced DE model, the  BE model does better than when it is compared to the benchmark model. 

\begin{table}
\begin{center}
\scalebox{0.55}{
\begin{tabular}{|c|c|c|c|c|}
\hline \bf Model Id & \bf Description & \bf Recall@1 & \bf Recall@2 & \bf Recall@5 \\ \hline
\bf DE & Dual Encoder (Benchmark) & $55.2$ & $72.09$ &  $92.43$\\
\hline
\bf DER & Dual Encoder Reproduced & $52.6 $  & $70.09 $ &  $91.51$\\
\hline
\bf BE & Bi-Encoder (Proposed) & $\bm{56.0}$ & $\bm{73.15}$ &  $\bm{92.7}$\\
\hline
\end{tabular}}
\end{center}
\caption{\label{BE-PER-table} Comparison of top-k $\%$ accuracy on UDCv2 on the test set}
\end{table}

Table~\ref{BE-EXP-table} shows the results of various experiments we performed on the BE model. 

\begin{table}
\begin{center}
\scalebox{0.55}{
\begin{tabular}{|c|c|c|c|c|}
\hline \bf Model Id &\bf Description & \bf Recall@1 & \bf Recall@2 & \bf Recall@5 \\
\hline
\bf BE-19 & BE with Cosine similarity & $43.09$  & $61.99$  & $86.97$\\
\hline
\bf BE-20 & BE with Polynomial Similarity & $55.11$ & $71.64$ &  $92.17$\\
\hline
\bf BE-21 & BE using all hidden states & $54.7$ & $71.54$ &  $91.63$\\
\hline
\bf BE-22 & BE with deep LSTM model & $53.8$ & $71.6$ &  $92.4$\\
\hline
\bf BE & BE with Dot Similarity & $\bm{56.88}$  & $\bm{73.24}$  & $\bm{92.86}$\\
\hline
\end{tabular}}
\end{center}
\caption{\label{BE-EXP-table} Results of different similarity measures used on the BE model using the validation set}
\end{table}

\begin{figure}
    \centering
    \includegraphics[scale=0.35]{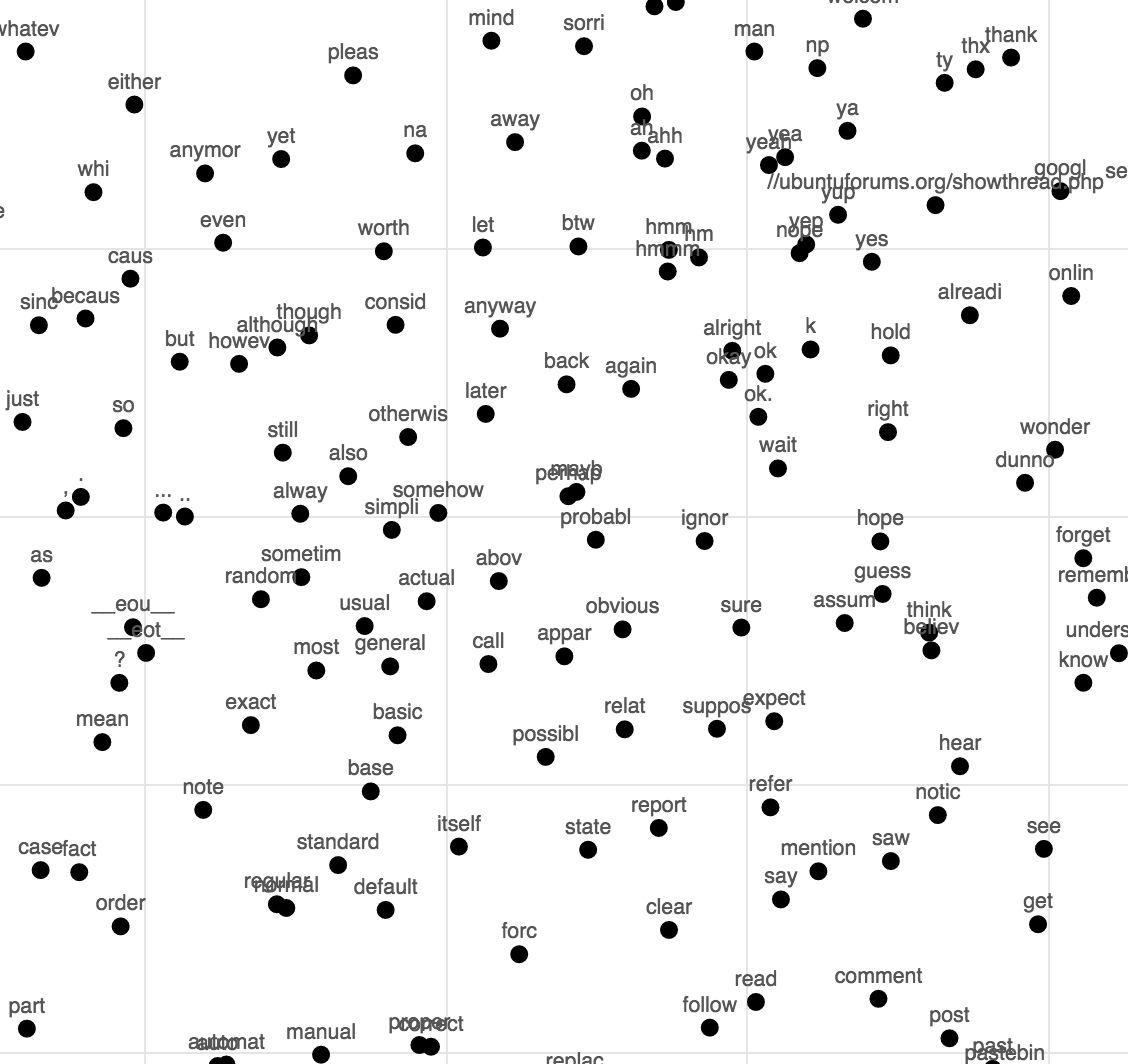}
    \caption{T-SNE plot of word embeddings of some frequently occurring words in UDCv2}
    \label{fig:tsne}
\end{figure}

 For a given  NLP task, choice of words embedding to real vector space can affect the performance of a model. Table~\ref{Embed-table} shows the results of using various embedding vectors with the BE model.We first looked at the random embedding and then used the Word2Vec embedding trained on the UDCv2. We also used the pre-trained GloVe embeddding \cite{Tomas:13} and ran the model with all four pre-trained GloVe embeddings that are available - (1) Wikipedia - 6B tokens, 400K vocab, uncased, 50d, 100d, 200d, and 300d vectors, (2) Common Crawl - 42B tokens, 1.9M vocab, uncased, 300d vectors, (3) Common Crawl - 840B tokens, 2.2M vocab, cased, 300d vectors, and (4) Twitter - 2B tweets, 27B tokens, 1.2M vocab, uncased, 25d, 50d, 100d and  200d vectors. Both pre-trained and trained embeddings on UDCv2 show better results than the random embedding. Between the Word2Vec and GloVe, the Common Crawl 42B embedding of the GloVe shows the best result. The T-SNE plot of Common Crawl 42B embeddings is shown in Figure~\ref{fig:tsne}. As can be seen in the diagram, similar words (for example - ``thank'', ``thx'' and ``ty'') appear embedded close to each other.

\begin{table}
\begin{center}
\scalebox{0.7}{
\begin{tabular}{|c|c|c|c|}
\hline \bf Embedding & \bf Recall@1 & \bf Recall@2 & \bf Recall@5 \\
\hline
\bf Random & $41.7$ & $61.1$ &  $87.8$\\
\hline
\bf Word2Vec & $56.55$ & $73.61$ &  $92.7$\\
\hline
\bf Twitter 27B 200d & $52.50 $ & $69.59$ & $91.44$ \\
\hline
\bf Common Crawl 42B & $56.88$ & $73.24$ & $92.86$ \\
\hline
\bf Common Crawl 840B & $56.43$ & $73.25$ & $92.66$\\
\hline
\end{tabular}}
\end{center}
\caption{\label{Embed-table} Comparison of performances of the BE model with various embedding types. Results are shown on the validation set}
\end{table}

\begin{figure}
    \centering
    \includegraphics[scale=0.6]{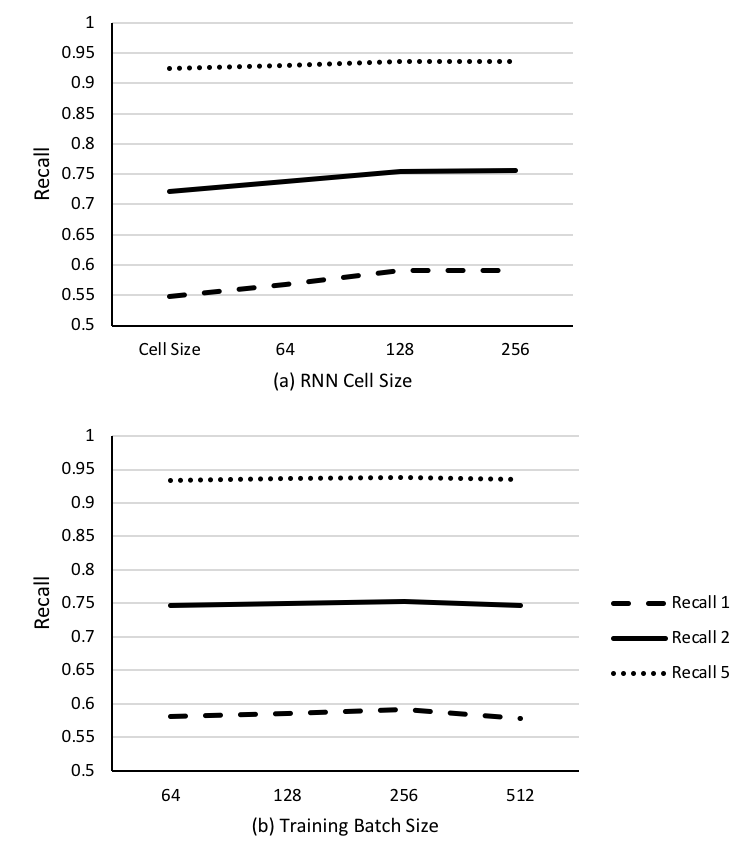}
    \caption{Effect of (a) RNN cell size and (b) training batch size on the BE (Bi-Encoder) model}
    \label{fig:cell_batch}
\end{figure}

In our experiments, we tuned LSTM cell size and the training batch size (Figure~\ref{fig:cell_batch}). 

\subsection{Error Analysis}
Similar to \newcite{Ryan:17}, we performed qualitative error analysis on the 100 randomly chosen examples from the test dataset where the model made an error for Recall@1 (Table~\ref{Error-table}). The errored examples were evaluated by three persons where each one manually gave a score to each examples for the metrics - Difficulty Rating, Model Response Rating and Error Category.

Difficulty Rating[1-5] measures how difficult human finds the context to match the right response. A rating of 1 on the difficulty scale means that the question is easily answerable by all humans. A 2 indicates moderate difficulty, which should still be answerable by all humans but only if they are paying attention. A 3 means that the question is fairly challenging, and may either require some familiarity with Ubuntu or the human respondent paying very close attention to answer correctly. A 4 is very hard, usually meaning that there are other responses that are nearly as good as the true response; many humans would be unable to answer questions of difficulty 4 correctly. A 5 means that the question is effectively impossible: either the true response is completely unrelated to the context, or it is very short and generic

Model Response Rating[1-3] measures the reasoning of the model's choice. A score of 1 indicates that the model predicted response is completely unreasonable given the context. A 2 means that the response chosen was somewhat reasonable, and that it’s possible for a human to make a similar mistake. A 3 means that the model’s response was more suited to the context than the actual response.

Error Category[1-4] puts model error in a specific category. Error Category of 1 relates to tone and style of the context. If a model makes an error attributed to the misspellings, incorrect grammar, use of emoticons, use of technical jargon or commands etc in the context, then the error category will be 1. Error Category 2 relates to when the context and chosen responses relate to the same topic. Error Category 3 relates to model's inability to account for turn-taking structure. For example if the last turn in the context asks a question and the model chose a answer where it is not answering the question. Error Category 4 means the model picked the response because it sees some common words between the context and the responses.

\begin{table}[H]
\begin{center}
\scalebox{0.8}{
\begin{tabular}{|c|c|}
\hline 
\bf Difficulty Rating & \bf \% of Errors \\
\hline
Impossible (5) & $19\%$ \\
Very Difficult (4) & $11\%$ \\
Difficult (3) & $22\%$ \\
Moderate (2) & $30\%$ \\
Easy (1) & $18\%$ \\
\hline
\bf Model Response Rating & \bf \% of Errors \\
\hline
Better than actual (3) & $23\%$ \\
Reasonable (2) & $21\%$ \\
Unreasonable (1) & $56\%$ \\
\hline
\bf Error Category & \bf \% of Errors \\
\hline
Common words (4) & $13\%$ \\
Turn-taking (3) & $45\%$ \\
Same topic (2) & $26\%$ \\
Tone and style (1) & $16\%$ \\
\hline
\end{tabular}}
\end{center}
\caption{\label{Error-table} Qualitative evaluation of the errors from the BE model}
\end{table}

The qualitative analysis results (Table~\ref{Error-table}) show that the BE model was not able to predict well the turn-taking structure of the dialogs. A little more than half of the errored examples had the human difficulty level ranging from 3 to 5, and almost half of the model responses in the errored examples were either reasonable or better than the actual response.

\section{Conclusions and Future Work}

This paper presented a new LSTM based RNN architecture that can score a set of pre-defined responses given a context(utterance). Empirically we have shown that on average 92.7\%, 73.15\% and 56.0\% of the time, correct response will
be in top 5, top 2 and top 1 correct responses respectively in Ubuntu Dialog Corpus Version 2 exceeding the accuracy of the benchmark model in all three metrics.
\newcite{Ronan:08} used a language model with a Rank loss/similarity where he had only positive
examples and generated negative examples by corrupting the positive ones. Several other works have
shown the Rank loss to be useful in training situations where pairs of correct or incorrect items are
to be scored \cite{Yoav:16}. Since UDC dataset matches this scenario, we recommend the
future work to explore the BE model with the Rank loss. In a large corpus like UDC where users are
seeking help in Ubuntu related problems, it is reasonable to assume that there can be multiple thread of
discussion(topics) related to Ubuntu. Identifying the latent topics and grouping the utterances based on
topics will allow training an ensemble of BE models. As there is no explicit grouping of the utterance,
we plan to identity these hidden topics using Latent Dirichlet Allocation (LDA). Topics distribution
of utterances can be used to group them using probabilistic measure of distance. We hypothesize that
ensembles of BE models may serve in efficient selection of correct responses.
Since the retrieval-based systems have to loop through every single possible responses, if the system
needs to go through a very large set, the system may be practically not feasible in production. As shown
by \cite{Anjuli:16} one way to reduce the number of possible responses is through clustering. \cite{Sina:10} and \cite{Hao:13} also showed several ways of reducing the large set of possible responses to a smaller set. We intend to apply such ideas in our future work.
Moreover, in a multi-turn dialog system capturing longer term context is essential to selecting correct
response. Our proposed architecture can be extended to more hierarchical RNN layers, capturing longer
context. We plan to investigate this further in conjunction with paragraph vector \cite{Quoc:14}.





\bibliography{bi_encoder}
\bibliographystyle{acl_natbib}

\appendix

\section{Source Code}
The source code for this project can be found here - \\ \url{https://github.com/DiwanshuShekhar/bi_encoder_lstm}
\end{document}